# MTS-CycleGAN: An Adversarial-based Deep Mapping Learning Network for Multivariate Time Series Domain Adaptation Applied to the Ironmaking Industry

Cedric Schockaert[†]
Paul Wurth S.A.
Luxembourg, Luxembourg
cedric.schockaert@paulwurth.com

Henri Hoyez
Institut supérieur de l'électronique
et du numérique
Lille, France
henri.hoyez@isen.yncrea.fr

## ABSTRACT

In the current era, an increasing number of machine learning models is generated for the automation of industrial processes. To that end, machine learning models are trained using historical data of each single asset leading to the development of asset-based dedicated models. To elevate machine learning models to a higher level of learning capability, domain adaptation, a category of transfer learning, has opened the door for extracting relevant patterns from several assets combined together. In this research we are focusing on translating the specific asset-based historical data (source domain) into data corresponding to one reference asset (target domain), leading to the creation of a multi-assets global dataset required for training domain invariant generic machine learning models. This research is conducted to apply domain adaptation to the ironmaking industry, and particularly for the creation of a domain invariant dataset by gathering data from different blast furnaces. The blast furnace data is characterized by multivariate time series. Domain adaptation for multivariate time series data hasn't been covered extensively in the literature. We propose MTS-CycleGAN, an algorithm for Multivariate Time Series data based on CycleGAN, a popular deep generative architecture for unsupervised translation of images from one domain to another. To the best of our knowledge, this is the first time CycleGAN is applied on multivariate time series data. Our contribution is the integration in the CycleGAN architecture of a Long Short-Term Memory (LSTM)-based AutoEncoder (AE) for the generator and a stacked LSTM-based discriminator, together with dedicated extended features extraction mechanisms. For this research, MTS-CycleGAN is validated using two artificial datasets embedding the complex temporal relations between variables reflecting the blast furnace process. MTS-CycleGAN is successfully learning the mapping between both artificial multivariate time series datasets, allowing an efficient translation from a source to a target artificial blast furnace dataset.

## CCS CONCEPTS

• Computing methodologies~Machine learning~Machine learning approaches~Neural networks

## 1 Introduction and background

With the tremendous amount of data collected, we are living in an era where machine learning model development will become the core activity for a lot of industries to automatize processes, to give insights about impact of decision on the company's business, or to give recommendations for future actions depending on the current state of an asset. The ironmaking industry has the challenging objective to bring the process control to a certain level of autonomy to optimize production, provide the workers a safer environment, control ecological impact of the production, optimally control the production of iron by using machine learning models predicting the hot metal temperature [1] taking into account material characteristics, environmental parameters, production target measured by key performance indicators (KPIs). The blast furnace is nowadays equipped with several thousands of sensors collecting enormous amount of data to measure temperatures, pressures, flows, chemical contents, etc. This data is employed to train a machine learning model to predict the temperature evolution of the hot metal produced by the blast furnace. The objective of such a model is to provide to the blast furnace operator insight about the evolution of the temperature allowing him to take the correct control decision to reach the production target. Predicting the hot metal temperature is made possible as a blast furnace has a high inertia dictated by the physicochemical underlying process and the size of the furnace itself. Charged materials (coke and iron ore) have a typical transfer time between 6 to 8 hours depending on mechanical design of the furnace and its actuators controlled by the process engineer and defining the operational mode of the blast furnace.

Each blast furnace having its own characteristics defined by its architecture, equipment and process operations, a machine learning model would require specific historical data for each blast furnace to train a model forecasting the hot metal temperature evolution. This is a limitation in the learning process of data-driven model requiring long historical data to cover only the dedicated operations of the corresponding blast furnace, without generalization that could be learned from other blast furnaces. An operation not learned by the machine learning model from the historical data of one blast furnace, will invalidate that model although that operation has potentially been already learned on another blast furnace by another machine learning model. There is as a clear need to apply technics known as transfer learning allowing to transfer the knowledge gathered by machine learning models trained on other blast furnaces, to a blast furnace having limited historical data. Transfer learning [2] is a field in machine learning that is gaining huge interests in various sectors. For industrial processes, each asset having a specific signature or operational mode, the machine learning models must be transferable from one asset to another one in order to optimally use the available data. Transfer learning is a key requirement for industry 4.0, in order to scale in the deployment of machine learning models.

In order to categorize transfer learning methods, two terms are defined in [2]: *domain* and *task*. A domain consists of a feature space and a marginal probability distribution (distribution of the features in the historical data), while a task consists of a label space and an objective predictive function. To illustrate this in the context of a model predicting the hot metal temperature, for a fix domain, a change in the feature space is resulting from the definition of features elaborated by the data scientist together with the process engineer that can be different between two furnaces (equipped with different sensors for example). A change of marginal probability distribution may result from a change in the operating mode of a blast furnace. On the other hand, for a task, a modification in the label space could for example be a modification of the classes to predict by a machine learning model between two furnaces, depending on the customer requirements. As an example, those classes could be: temperature range forecasting, temperature tendency (increasing more than 5°C, decreasing more than 5°C, stable in a range [-5°C, +5°C]), specific phenomena (sudden drop of temperature, sudden increase of temperature, deviation from target by $x$%), etc. A change in the objective predictive function may results from a change in the label distribution from the historical data of two different blast furnaces, respectively the *source* and the *target* blast furnace.

Three terms are introduced in [2] to classify transfer learning algorithms: *inductives*, *transductive* and *unsupervised*. Inductive transfer learning is characterized by a difference of tasks between the source and target domains while both domains are similar. An additional requirement is the availability of labels in the target domain. In transductive transfer learning, the domain is changing between the target and the source, while the tasks remain the same. Another requirement for transductive transfer learning is that labels from the source are potentially available but not in the target domain. Finally, unsupervised transfer learning doesn't require any labels from the source and target domains, and tasks are differing similarly to inductive transfer learning.

Domain adaptation [3], is part to the group of transductive transfer learning. Both tasks in the source and target domains remain the same, however the domain differs. The only assumption is that the data is coming from both the source and target distinct domains. If labels are associated to the data, authors are mentioning supervised domain adaptation. Similarly, semi-supervised and unsupervised domain adaptation are the terminology adopted in the literature for the transferability between domains of respectively semi-supervised and unsupervised machine learning models. As an illustration of supervised domain adaptation, a model predicting the hot metal temperature trained with the data of one blast furnace (source domain) is transferred to another blast furnace (target domain) without using labels (temperature measurements) from that blast furnace.

In this paper we will focus on domain mapping. The objective is not to transfer an existing model between the source and the target domains, so the term task is not relevant for this research, but to learn the mapping function allowing a translation of the target domain data into source domain data. By working on learning a data mapping between domains, a centralized domain invariant dataset can be created allowing more efficient comparison of assets for the expert, and is a trigger to train generic models gathering the 'experience' of each individual asset, and therefore to increase the forecasting robustness of derived data-driven predictive models. Another approach in the literature is to transform existing features for a machine learning model into domain invariant features [4]. As a results, a model trained on those features can be applied either to the source or target data. However, this is not a solution in line with our requirement to build a centralized domain invariant dataset.

Domain mapping is typically created adversarially. Adversarial training refers to methods that utilize an adversarial process during the training [5, 6]. Adversarial training is characterized by putting two neural networks against each other, playing the role of a data discriminator and data generator. The generator tries to generate data of the source domain from data of the target domain, while the discriminator is attempting to make the difference between real data from source domain and translated data from target to source domain. Both networks are playing a minimax game during the training where the generator tries to fool the discriminator, while the discriminator attempts not to be fooled. Those architectures based on Generative Adversarial Networks (GANs) [7], are facing several challenges for training such as difficulty of converging [8], mode collapsing where the generator is learning to generate only artificial samples from few specialized modes of the data distribution [9], and the usual vanishing gradient of deep learning [7]. Nowadays, GAN-based architectures have been applied mostly for synthetic image domain adaptation [10, 11, 12, 13]. CycleGAN [13], and some of its variants [14, 15], is a popular architecture for unsupervised domain adaptation for images. As an example of application, CycleGAN learns to generate night vision images from day vision images, without having corresponding pair of images from both domains of the same scene for the training (unsupervised).

The application of domain adaptation for time series data hasn't been extensively researched. Few solutions have been proposed like VRADA [16] where temporal features of healthcare datasets are extracted by means of a Variational Recurrent Neural Network (VRNN) trained adversarially, or using Long Short-Term memory (LSTM) [17] rather than a Convolutional Neural Network (CNN)

MTS-CycleGAN: An Adversarial-based Deep Mapping Learning Network for Multivariate Time Series Domain Adaptation Applied to the Ironmaking Industry for images. In [18] a combination of CNNs and Recurrent Neural Networks (RNNs) is used to identify sleep stages from radio spectrogram temporal modifications. Other applications are described in the literature for speech recognition [19] using bi-LSTMs, or text classification [20] where LSTMs are implemented for features extraction.

In this paper, we propose MTS-CycleGAN, an unsupervised domain adaptation architecture for Multivariate Time Series based on CycleGAN. Our contribution is the development of a generator and discriminator for CycleGAN that are dedicated for multivariate time series by implementing respectively a stacked LSTM-based AutoEcoder (AE) and a stacked LSTM-based binary classifier including dedicated extended features extraction mechanisms. MTS-CycleGAN is trained and validated using two datasets created artificially and representing a source and target blast furnace. Each artificial dataset is embedding the characteristics of blast furnace data where each variable has a different reaction time on other variables due to the high inertia of the underlying process, and where there are amplitude changes between signals of both furnaces reflecting potential difference of size or operation mode. Two artificial datasets are generated to represent the source and the target domains respectively. The validation of MTS-CycleGAN is achieved using the prior knowledge of both domains being the parameters used to generate both artificial datasets characterized by different temporal shifts and signal amplitudes.

In the next section, the deep learning architecture of the proposed approach is described. Results are presented on an artificial datasets. Conclusion and perspectives of this research are discussed.

## 2 Description of the proposed approach and results

The CycleGAN architecture is schematized in Figure 1a. $\mathbf{X_s}$ and $\mathbf{X_t}$ denote respectively two unpaired multivariate time series datasets from the source domain, and the target domain. $x_s$ and $x_t$ denote two samples respectively from $\mathbf{X_s}$ and $\mathbf{X_t}$. The CycleGAN architecture is characterized by two generators $G_{st}$ and $G_{ts}$ respectively for source-to-target and target-to-source data transformation, and two associated discriminators $D_s$ and $D_t$. The discriminator $D_t$ (resp. $D_s$) encourages $G_{st}$ (resp. $G_{ts}$) to generate fake data $x^g_t$ (resp. $x^g_s$) that cannot be distinguished from the real target (resp. source) domain, known as the adversarial loss. To further regularize the mappings, additional losses are introduced: forward/backwards cycle consistencies and the identity loss (see Figure 1a). The global loss $L_{g,st}$ (resp. $L_{g,ts}$) is a weighted sum of those 4 losses to train the generator $G_{st}$ (resp. $G_{ts}$). Each generator or discriminator of the architecture is trained sequentially while others are not trainable. This sequence is repeated until convergence of the training.

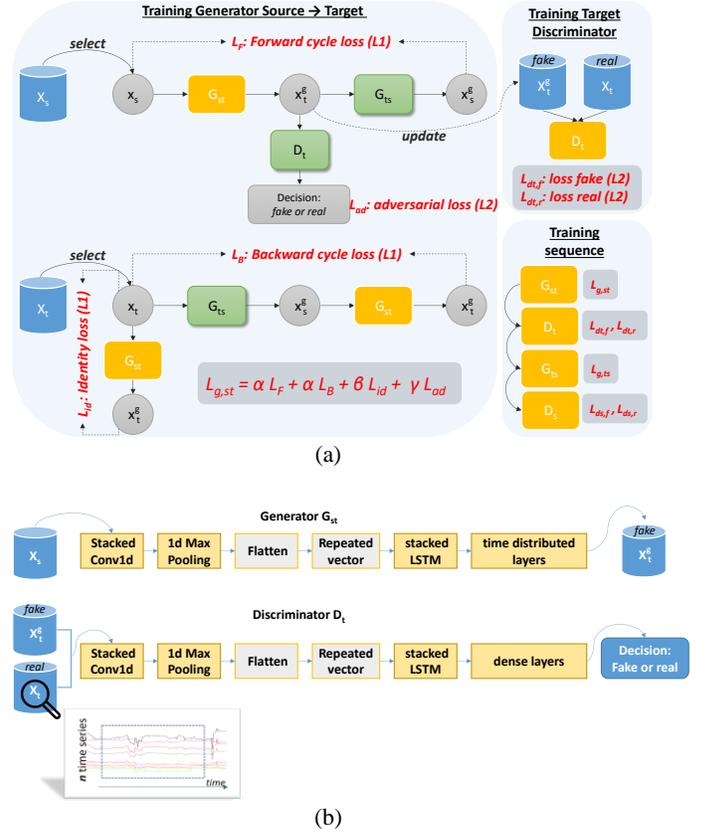

(a)

(b)

**Figure 1: (a) MTS-CycleGAN architecture and training procedure; (b) Generator and Discriminator architectures for multivariate time series**

A dedicated architecture of the generators and discriminators for multivariate time series is described in Figure 1b. It is composed of stacked convolutional and LSTM layers for features extraction. The generator is reconstructing multivariate time series while the discriminator is classifying multivariate time series within a temporal window as fake (generated by the generator), or real.

Artificial unpaired multivariate time series datasets for the source and target domains are generated to train and validate MTS-CycleGAN, as illustrated in Figure 2. Those datasets are characterized by different temporal shifts and amplitude. Those features have been carefully defined as they are representing the complexity inherent to the blast furnace signals. Three parameters are introduced for the generation of those datasets: $\alpha$, $\beta$ and $\gamma$. Repeated amplitude modification on $A(t)$ by a random value in a range [5, 50] and triggered at a random time in a range [0, 3h] in a time window of 6h, is impacting other time series with a temporal shift and a change of amplitude.

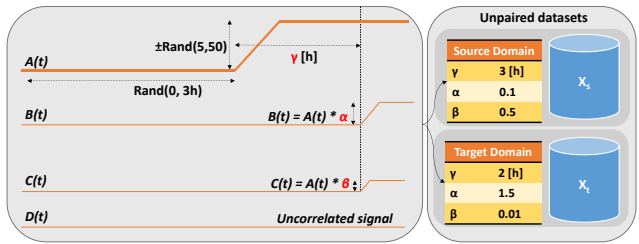

**Figure 2: unpaired multivariate time series artificial datasets**

MTS-CycleGAN has been trained on two artificial datasets corresponding to multivariate time series of a source and target domains. A temporal window size of 6h is used and defines a sequence of four time series.

Results are presented in Figure 3 where a source and target sequence is mapped to the target and source domains respectively, as an illustration. The retrieved values for the parameters $\alpha$, $\beta$ and $\gamma$ are analyzed for more than 200 sequences per domain after domain adaptation by applying MTS-CycleGAN. The median values calculated for those parameters can be compared with the ones used to generate the dataset for each domain.

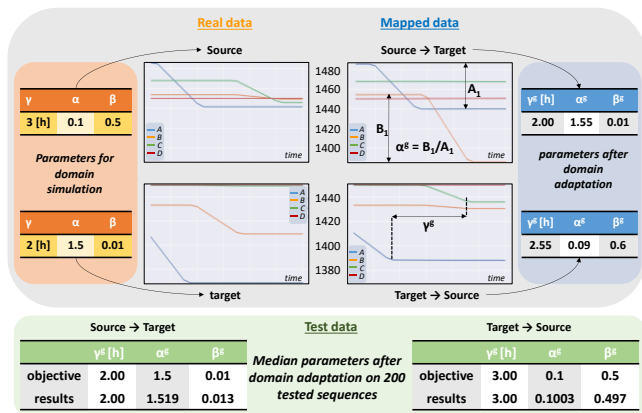

**Figure 3: MTS-CycleGAN results to retrieve parameters of source and target artificial datasets after domain adaptation**

## 3 Conclusion and perspectives

MTS-CycleGAN, a domain adaptation architecture for multivariate time series based on CycleGAN, has been trained and validated on a source and target artificial dataset embedding characteristics of signals recorded on a blast furnace. The results show it is achievable to retrieve the parameters that define the source and target domains, from the other domain after mapping by applying MTS-CycleGAN. An average median error of 8% has been reached for the parameters $\alpha$ and $\beta$ influencing the amplitude of one time series, while the median error for the temporal shift is perfectly matching the parameter $\gamma$ defined for the creation of the artificial dataset for each domain.

With the results obtained in this research, we are confident to apply MTS-CycleGAN to learn the mapping between two different blast furnaces, although several challenges still need to be addressed being the large number of sensors, the amount of data required for training this architecture and the validation of the results by a process engineer acting like a discriminator to accept or reject a blast furnace dataset generated by MTS-CycleGAN.